\documentclass[10pt,twocolumn,letterpaper]{article}

\usepackage{style/cvpr}
\usepackage{times}
\usepackage{epsfig}
\usepackage{graphicx}
\usepackage{amsmath}
\usepackage{amssymb}
\usepackage{float}
\usepackage{subfig}

\belowdisplayskip \abovedisplayskip
\newlength{\sectionReduceTop}
\newlength{\sectionReduceBot}
\newlength{\subsectionReduceTop}
\newlength{\subsectionReduceBot}
\newlength{\abstractReduceTop}
\newlength{\abstractReduceBot}
\newlength{\captionReduceTop}
\newlength{\captionReduceBot}
\newlength{\subsubsectionReduceTop}
\newlength{\subsubsectionReduceBot}

\newlength{\eqnReduceTop}
\newlength{\eqnReduceBot}

\newlength{\horSkip}
\newlength{\verSkip}

\newlength{\figureHeight}
\newcommand{\csection}[1]{\vspace{\sectionReduceTop}\section{#1}\vspace{\sectionReduceBot}}
\newcommand{\csubsection}[1]{\vspace{\subsectionReduceTop}\subsection{#1}\vspace{\subsectionReduceBot}}

\newcommand{\cabstract}[1]{\begin{abstract}\vspace{\abstractReduceTop}#1\end{abstract}\vspace{\abstractReduceBot}}

\usepackage[breaklinks=true,bookmarks=false]{hyperref}

\cvprfinalcopy % *** Uncomment this line for the final submission

\def\cvprPaperID{****} % *** Enter the CVPR Paper ID here

\newcommand{\xhdr}[1]{\vspace{2pt}\noindent\textbf{#1}}

\begin{document}

%%%%%%%%% TITLE
\title{Question-Conditioned Counterfactual Image Generation for VQA}

\author{Jingjing Pan \thanks{Carnegie Mellon University ~~~~ $\dagger$ Georgia Tech} \\
{\tt \small jingjinp@andrew.cmu.edu}
\and Yash Goyal$^\dagger$\\
{\tt \small ygoyal@gatech.edu}
\and Stefan Lee\footnotemark[2]\\
{\tt \small  steflee@gatech.edu}}
\maketitle

\cabstract{
    %Visual Question Answering (VQA)\cite{vqa} remains a challenging problem: Why does the VQA algorithm predict this answer? What regions of the image does it pay more attention to given a question? To understand and improve current VQA models, we develop a counterfactual image generation algorithm to build negative VQA examples. Formally, given a VQA triplet (image $I$, question $Q$, answer $A$), our goal is to generate a counterfactual image $I'$ that has a slight amount of difference from the original image $I$, such that the original answer $A$ does not apply to question $Q$ any more. By investigating such negative examples, we can suppose that the most different region between $I$ and $I'$ is the most important factor that influences the answer produced by the VQA algorithm. In the future, we may expect improvement in VQA network by training with such negative examples.
    %
    While Visual Question Answering (VQA) \cite{vqa} models continue to push the state-of-the-art forward, they largely remain black-boxes -- failing to provide insight into how or why an answer is generated. In this ongoing work, we propose addressing this shortcoming by learning to generate counterfactual images for a VQA model -- \ie given a question-image pair, we wish to generate a new image such that i) the VQA model outputs a different answer, ii) the new image is minimally different from the original, and iii) the new image is realistic. Our hope is that providing such counterfactual examples allows users to investigate and understand the VQA model's internal mechanisms.
}

\csection{Introduction}

While VQA models have steadily improved over the years, they are still prone to making somewhat silly errors that can leave human users baffled. In these situations, a user might ask for an explanation as to why the model answered as it did rather than with some alternative response for a given image and question. One way to respond to this request for discriminative explanation is through counterfactuals -- that is, to present the user with other images for which the model produces the requested alternative response given the original question. But to be most instructive, these images ought to be as similar to the original as possible. In this ongoing work, we explore learning to generate such question-conditioned counterfactual images.

\xhdr{Counterfactual Visual Explanations.} While there is extensive work on generating explanations for deep models \cite{ribeiro,doshi,julius}, only a few works have addressed counterfactual based visual explanations \cite{yash,openset}. Unlike our work which generates images, \cite{yash} and \cite{openset} operate in feature rather than pixel space and apply to the context of image classification rather than language-conditioned settings.
% On the other hand, \cite{explain} operates in pixel space but focuses on building counterfactuals by removing image regions and replacing them with generated patches. These approaches have been applied in the context of image classification but not in language-conditioned settings.

\xhdr{Language-Conditioned Image Generation.} Critically, counterfactual explanations for VQA must be conditioned on the question, \eg while changing the color of a skateboard wheel is an excellent counterfactual for ``What color is the wheel?'' the same edit would be irrelevant to a question about the board's deck. As such, recent work on generating images based on natural language captions \cite{attngan} or dialogs \cite{chatpainter} about the image is closely related. However, our setting lacks explicit target images and must generate images such that a VQA model changes its decision.

\csection{Method}
\begin{figure}
  \centering\includegraphics[width=1\linewidth,height=1.25in]{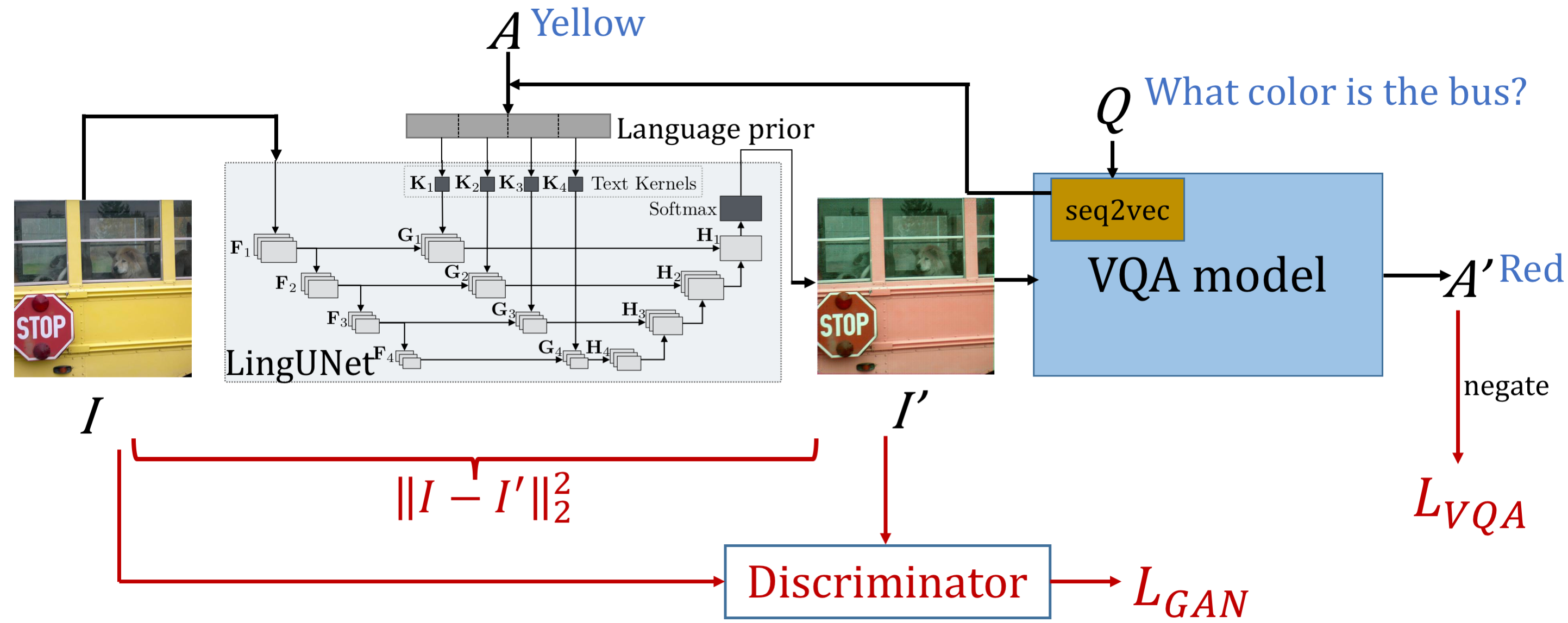}\\[-8pt]
  \caption{We learn to generate counterfactual visual explanations for VQA models. For example, to explain why a model predicted the bus to be in yellow color in the example above, we want to generate a new image $I'$ that is similar to the original but results in a different answer (white in this case) so the user can see what the most important factors are in the decision\protect\footnotemark.}
  \label{fig:model}
  \vspace{-10pt}
\end{figure}

\footnotetext{The way VQA v2.0 \cite{vqav2} was collected shares the `minimal edit' notion, where human were asked to pick nearest neighbor images s.t. one question can be applied to multiple images with minimal difference.}

Given a VQA model $f\colon (I,Q) \rightarrow \hat{A}$ and a dataset of image-question-answer tuples $\{(I_i,Q_i,A_i)\}_{i=1}^N$, our goal is to learn a model $g\colon (I,Q,A) \rightarrow I'$ that observes the image $I_i$, question $Q_i$, and ground-truth answer $A_i$ and then generates a counterfactual image $I_i'$ such that the answer for the new image is different than that for the old, \ie $f(I,Q) \neq f(I',Q)$. Further, we require the differences between $I_i$ and $I_i'$ to be minimal and for $I_i'$ to be visually realistic.

\xhdr{Language-Conditioned Image Editing.} The general architecture of our model is shown in Fig.~\ref{fig:model}. We instantiate the counterfactual generator $g\colon (I,Q,A) \rightarrow I'$ as a LingUNet \cite{lingunet} architecture that maps conditioning language to key intermediate filter weights in the popular pixel-to-pixel UNet model \cite{unet}. For language conditioning, we encode the question based on the VQA model's language encoding and the final logit weight vector for the VQA model's answer. These are concatenated and passed through a fully-connected layer before being passed to LingUNet.

\xhdr{Constraining Image Generation.} We train the counterfactual generator under three losses, each corresponding to one of our desired traits. First, the generated image should change the VQA model's response so we train with the negated cross-entropy from the VQA model with the original answer $A$ as the target, \ie we train for any answer \emph{other than $A$}. Second, we want edits to be minimal so we add an $\ell_2$ loss between the original and generated image, \ie $||I, I'||_2$. Finally, we introduce a discriminator (as in GAN training \cite{goodfellow2014generative}) that penalizes unrealistic generated images.

\csection{Experiments \& Results}
%\vspace{5pt}
%\csubsection{Experiment Details}
%We use the VQA dataset \cite{vqa} based on MS COCO\cite{COCO} images for evaluation. This is a large dataset that contains 204,721 images, each with 3 question-answer pairs. Images are resized to 448 by 448 for our pipeline.

%We start with MLB model pre-trained on VQA dataset. We warmup LingUNet by $\ell_2$ reconstruction loss using original images. We use the 4800-D question feature extracted by skip-thoughts\cite{skip-thought} model inside pre-trained in MLB. For answer feature, we directly take the weights (2400-D for each answer) from the last fully-connected layer in pre-trained MLB. Then, we train our model end-to-end with $\mathcal{L}_{model}$ for 1 epoch (TRAINING FOR MORE EPOCHS) until the open-end accuracy drops from 64.33\% to 59.87\% for all questions and from 72.64\% to 38.38\% for color-related questions. Here we the "what color is" question type as color-related questions. We choose $\lambda = 0.03$. Adam\cite{adam} optimizer is used with 0.001 as learning rate and 0.1 as gradient clipping.

%We train our model $g$ end-to-end with $\mathcal{L}_{model}$ and $\mathcal{L}_{GAN}$ alternatively. For the discriminator $D$,  We follow a similar architecture with DCGAN\cite{dcgan}. The optimizer for $D$ is Adam with 0.0001 as learning rate.  Inspired by the label smoothing technique used in \cite{softlabel}, we use 0.9 as real label and 0.1 as fake label to reduce the vulnerability of $D$ network to adversarial examples.

\xhdr{Dataset.} We apply our approach to the VQAv1 \cite{vqa} dataset and use a pre-trained MLB model \cite{Kim2017} as the VQA model. 

\xhdr{Training.} We warm up LingUNet using only the $\ell_2$ loss for reconstruction before adding in the question-answer inputs. We then add question-answer conditioning and train with all three losses for 6 epochs. We use the Adam optimizer, gradient clipping, and soft-labels for the discriminator \cite{softlabel}. 

%=========================================================================
\csubsection{Results}
%\begin{figure}[t]
%    \centering
%    \includegraphics[width=1\linewidth,height=2in]{figures/success.png}
%    \caption{Successful examples where our model generated counterfactual image conditioned on the question with minimal change.}
%    \label{fig:success}
%\end{figure}

We explore our counterfactual image generation algorithm under three criteria: (a) change of semantics, (b) sensitivity to language-conditioning, and (c) realism.

\begin{figure}[t]
    \captionsetup[subfloat]{farskip=1pt,captionskip=1pt}
    \centering
    \subfloat[]{\label{fig:hydrant}\includegraphics[width=1\linewidth,height=1.3in]{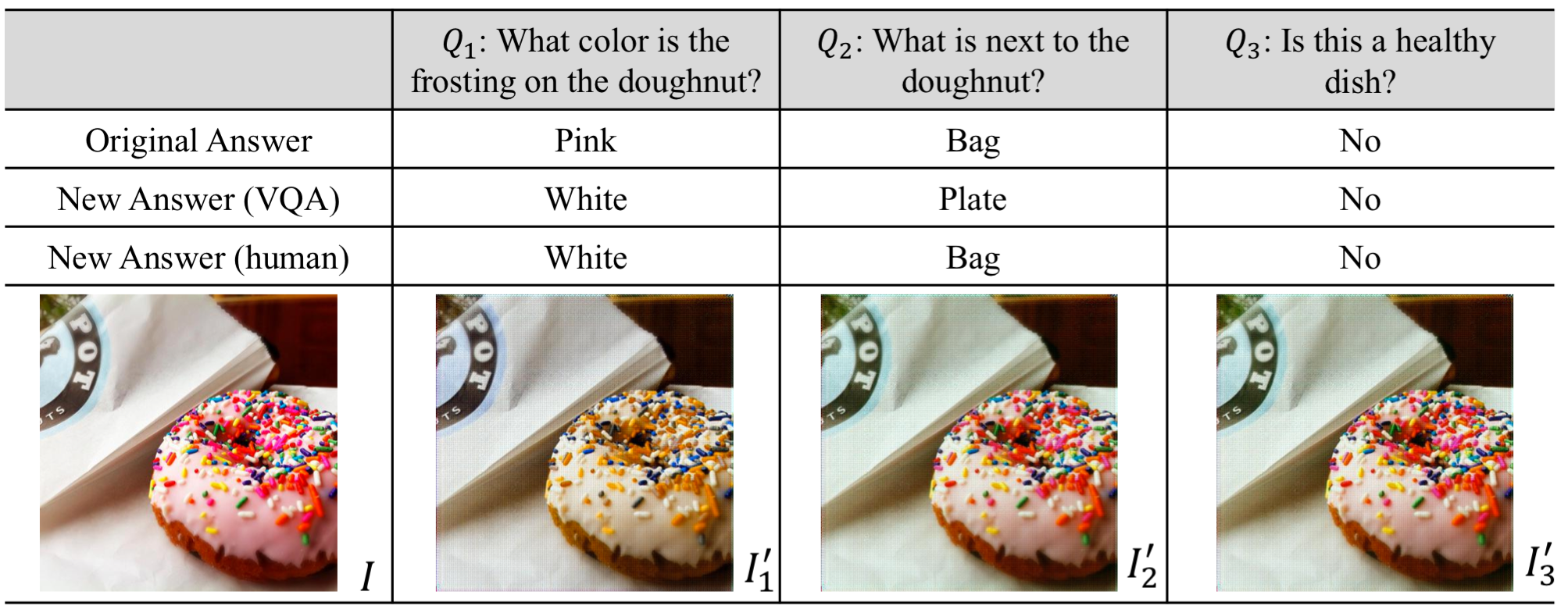}}\\
    \subfloat[]{\label{fig:skateboard}\includegraphics[width=1\linewidth,height=1.3in]{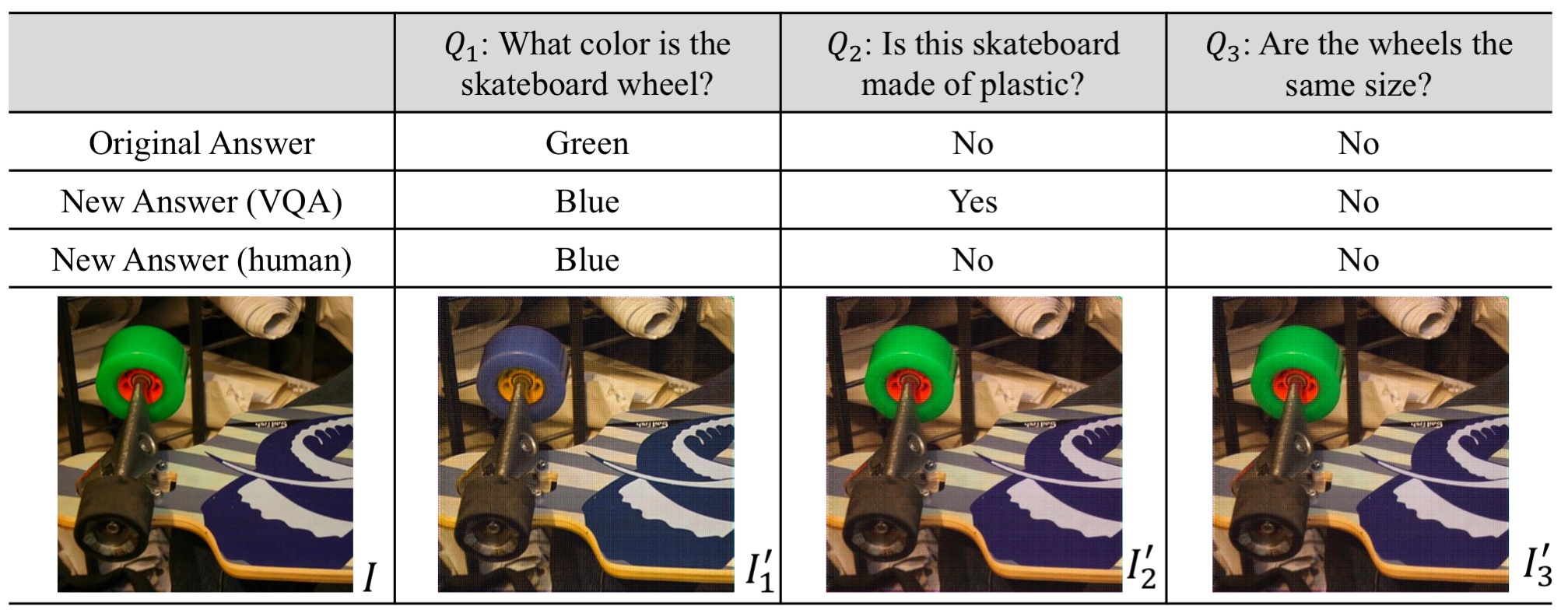}}\\
    \subfloat[]{\label{fig:kitchen}\includegraphics[width=1\linewidth,height=1.3in]{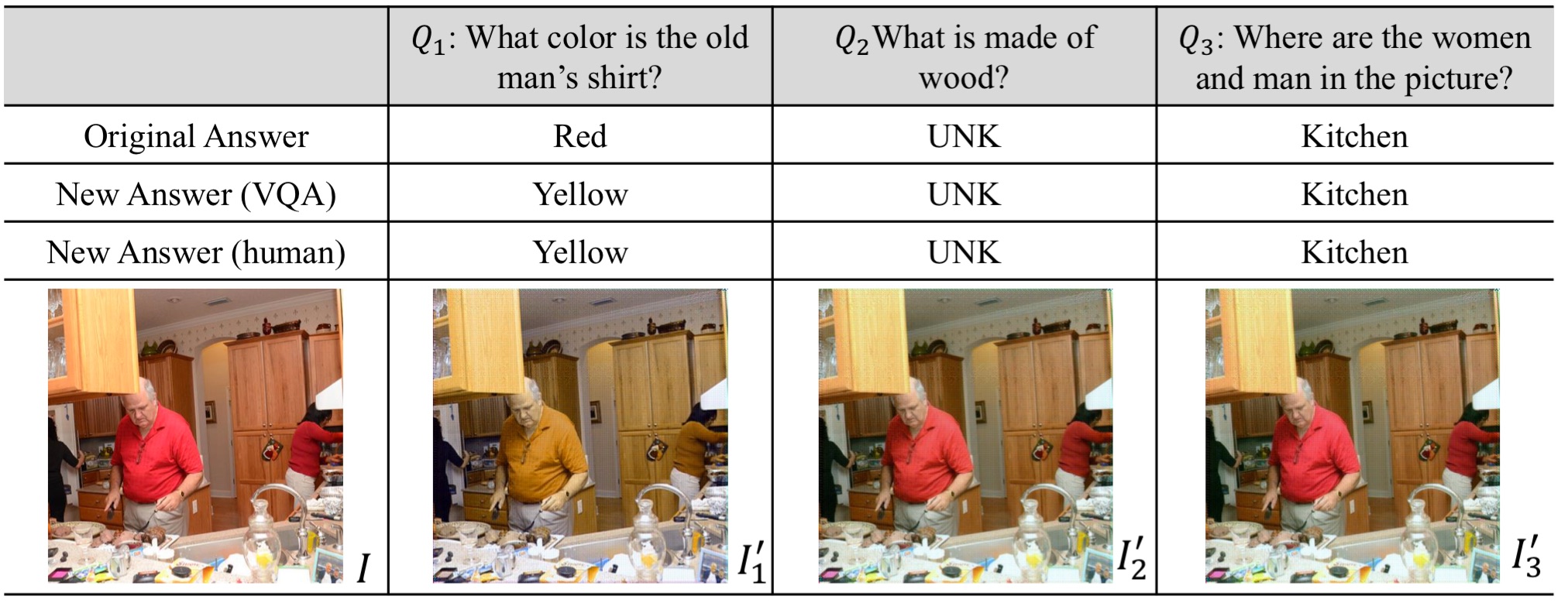}}
    \caption{Examples of generated images and answer to them given by VQA model and human.}
    \label{fig:change}
    \vspace{-10pt}
\end{figure}

\xhdr{Change of semantics.} Ideally, the counterfactual images generated will change the semantics of the image based on the question and original answer -- resulting in a different answer for both the VQA system and human observers. We find VQA accuracy on the validation set drops from 64.33\% to 57.64\% when using our generated examples. Moreover, we find a significantly larger effect for ``what color'' questions, 72.64\% to 34.02\%. This means that a large portion of generated counterfactual images have successfully changed semantics that are observable by the VQA system. Fig.~\ref{fig:change} shows some qualitative results of our approach. We can see that for color questions ($Q_1$) the image regions in question are shifting color to an alternative that both the VQA and human annotator can identify. However, for other question types ($Q_2$ and $Q_3$ in Fig.~\ref{fig:change}), our approach is either ineffective at changing the VQA model's answer or does so in a way that is not human perceptible, leading to disagreement as in $Q_2$ in Fig.~\ref{fig:skateboard}. Another failure mode is over-editing of irrelevant objects as in the woman's shirt for $Q_1$ in Fig.~\ref{fig:kitchen}. 

\xhdr{Sensitivity to language-conditioning.} As we can see from Fig.~\ref{fig:change}, the model does learn to produce different image edits for different input questions-answer pairs. However, the resulting edits for non-color questions don't seem to be modifying the image significantly.

%First, the algorithm should generate different images according to different questions. Our model satisfies this criterion under the restriction to color-related questions. As shown in Figure \ref{fig:change}, the images generated by our model from color-related questions ($Q_1$) are different from the other 2 generated images.

%\xhdr{Realism.} Last, generated images should have realistic appearance and contain as less artifacts as possible. In most cases, our model does not generate obvious artifacts (Figure \ref{fig:success}, \ref{fig:change}). However, there are also some cases when our model generated patterned artifacts (small dark spots in Figure \ref{fig:artifacts}) to change the VQA answer, but not the semantic information that human eyes can recognize. Although we train a discriminator along with $g$ to avoid $g$ generating artifacts, in practice it is known to be hard to balance the training process of discriminator and generator. In the future, we will explore different   GAN configurations to alleviate artifacts in $I'$.

\xhdr{Realism.} In most cases, our model does not generate obvious artifacts; however, there remain light checkerboard patterns in some output images (visible in Fig.~\ref{fig:hydrant}). Achieving even this level of realism required significant fine-tuning of the discriminator training regime to avoid the introduction of adversarial ``deep dream'' style artifacts like superimposing dog faces into the images to fool the VQA model.

%\begin{figure}[ht]
%    \centering
%    \includegraphics[width=1\linewidth,height=1.03in]{figures/artifact.png}
%    \caption{Failure examples when patterned artifacts appear in $I'$ to change the VQA answer.}
%    \label{fig:artifacts}
%\end{figure}
\csection{Discussion}
\label{sec:disc}

Our current experiments have shown our proposed model effective for language-conditioned counterfactual image generation in the context of color questions, but has yet to show promise with larger semantic edits (e.g. changing wheel size in Fig.~\ref{fig:skateboard}, $I_3^{'}$). This may be due to our $\ell_2$ loss that may be overly constraining for larger scale edits. We are exploring different configurations to train a better discriminator (i.e., converged to Nash Equilibrium) so that we can loosen $\ell_2$ constraint for larger semantic edits.

Further, our model seems to make color edits on all relevant areas (both the man and woman's shirts in Fig.~\ref{fig:kitchen}). Shifting to a model which predicts not only edits but a spatial mask for where to apply them may resolve both these issues -- localizing edits and allowing for a minimum edit loss to be applied to the mask rather than individual pixels. 

\clearpage

{\small
\bibliographystyle{style/ieee_fullname}
\bibliography{egbib}
}
\end{document}